\documentclass[10pt,twocolumn,letterpaper]{article}

\usepackage{cvpr}
\usepackage{times}
\usepackage{epsfig}
\usepackage{graphicx}
\usepackage{amsmath}
\usepackage{amssymb}

\usepackage[breaklinks=true,bookmarks=false]{hyperref}

\cvprfinalcopy 

\def\cvprPaperID{****} 

\begin{document}

\title{Team JL Solution to Google Landmark Recognition 2019}

\author{
\begin{tabular*}{0.7\textwidth}{@{\extracolsep{\fill}}cc}
Yinzheng Gu & Chuanpeng Li
\end{tabular*}\\
Jilian Technology Group (Video++)\\
{\tt\small \{guyinzheng, lichuanpeng\}@videopls.com}
}

\maketitle

\begin{abstract}
In this paper, we describe our solution to the Google Landmark Recognition 2019 Challenge held on Kaggle. Due to the large number of classes, noisy data, imbalanced class sizes, and the presence of a significant amount of distractors in the test set, our method is based mainly on retrieval techniques with both global and local CNN approaches. Our full pipeline, after ensembling the models and applying several steps of re-ranking strategies, scores 0.37606 GAP on the private leaderboard which won the 1st place in the competition.
\end{abstract}

\section{Introduction}\label{sec:intro}

To foster progress in landmark recognition, the Google-Landmarks-Dataset (GLD) \cite{NASWH2017} was released last year together with two competitions (Google Landmark Recognition and Retrieval Challenges) on Kaggle. For the recognition challenge, the training set consists of 1,225,029 images belonging to 14,951 classes whereas the test set consists of both landmark and non-landmark/distractor images for a total of 117,703 images. For each test image that depicts a landmark, one is asked to predict the correct landmark class together with a confidence score whereas for distractors one should leave an empty prediction in the submission. The evaluation metric for the competition is GAP, so that it is important to make sure distractors, if predicted, should have lower confidence scores than real landmark images.

This year, Google released the second version of the dataset known as Google-Landmarks-Dataset-v2 (GLD2) together with two new competitions (Google Landmark Recognition and Retrieval 2019 Challenges) on Kaggle. For the recognition challenge, same evaluation metric is used as last year, and the test set is of similar size with 117,577 images in total of both landmark and distractor images. The training set, on the other hand, is much larger with 4,132,914 images belonging to 203,094 classes. Moreover, unlike last year, the training set was released without any data cleaning step and hence is much more diverse. Due to these reasons, our final solution is comprised entirely of retrieval methods, unlike last year in which a non-trieval part of our winning solution was based on CNN classification models.

In the rest, we describe the full solution in details in Section \ref{sec:sol}, present the results in Section \ref{sec:results}, and conclude in Section \ref{sec:conclu}.

\section{Proposed Solution}\label{sec:sol}

We now describe our approach in details, which consists of the follwing models and main steps.

\subsection{Dataset Cleaning}\label{sec:cleaning}

As early exploratory data analysis on the provided training set indicates extreme class imbalance and visually unrelated images within the same class, we begin by data cleaning with the following simple process.

We first remove all classes with no more than 3 training samples (53,435 classes in total), then we take one of the best fine-tuned models from previous year's retrieval challenge and extract descriptors of all remaining images. In our case, the architecture consists of ResNeXt-101 (64$\times$4d) as backbone and GeM as the pooling operation (to be explained later), but one can certainly substitute with other choices/methods as long as the performance is satisfactory on the leaderboard. For each remaining class, we match all images within the class with each other using the extracted descriptors, and consider two images as a matching pair if the corresponding cosine similarity is above a certain threshold (0.5 in our case). After this process, 37,877 more classes are discarded as there are no matching pairs. For the rest, we take the maximum connected component of the tree for the class which is constructed by setting the remaining images as vertices and connecting two vertices by an edge if they form a matching pair. This leaves 836,964 images in total belonging to 112,782 classes from which we randomly select at most 100 pairs for each class resulting in 2,153,615 matching pairs. All models below are fine-tuned on this clean version of the training set.

\subsection{Global CNN models}\label{sec:globalmodels}

By a global CNN retrieval model we mean a model that takes an input image and produces an $\ell^2$-normalized compact image descriptor so that the cosine similarity between two images can be efficiently calculated by taking the dot product of the descriptors. In this competition, we choose the ``combination of multiple global descriptors'' (CGD) proposed in \cite{JKKKK2019} as our framework (with some modifications). This is also partially inspired by the fact that last year's winning solution of the retrieval challenge consists of a weighted concatenation of six descriptors. More specifically, we use ResNet-101 \cite{HZRS2016},  ResNeXt-101 (64$\times$4d) \cite{XGDTH2017}, SE-ResNet-101 \cite{HSS2018}, SE-ResNeXt-101 (32$\times$4d) \cite{HSS2018}, and SENet-154 \cite{HSS2018} as the backbone networks. For each backbone, we take the feature maps produced by the last convolutional layer and combine with the attention maps proposed in \cite{GLX2018}. Then we apply the following pooling operations: GeM \cite{RTC2018}, RMAC \cite{TSJ2016}, MAC \cite{TSJ2016}, and SPoC \cite{BL2015}, each of which produces an image descriptor of 2048-D, which is then appended by a fully connected layer (with bias) of output 1024-D. The resulting descriptors (4$\times$2048-D + 4$\times$1024-D) are concatenated together followed by $\ell^2$-normalization forming a compact image descriptor. For dimensionality reduction, we use attenuated unsupervised whitening (AUW) \cite{MTBJC2018} with $t = 0.5$ learned on the clean version of the training set.

During training, all images are resized before feeding to the network without distorting the original aspect ratio. Following \cite{RTC2018}, every epoch contains 2,000 tuples divided into batches of size 10, where each tuple is of the form $(I_q, I_p, I_{n, 1}, \ldots, I_{n, 5})$ with $I_q$ being a query image, $I_p$ being a positive image matching $I_q$, and $I_{n, 1}, \ldots, I_{n, 5}$ being hard negative non-matching images selected from a pool of 60,000 images. Positive images are fixed while negative images are re-mined every epoch. The loss function being minimized is the sum of a contrastive loss function (with margin 0.9) and a triplet loss function (with margin 0.2). For the contrastive loss, every tuple is converted to 6 pairs
\[(I_q, I_p), (I_q, I_{n, 1}), \ldots, (I_q, I_{n, 5}),\]
whereas for the triplet loss, every tuple represents 5 triplets
\[(I_q, I_p, I_{n, 1}), \ldots, (I_q, I_p, I_{n, 5}).\]
During test time, we extract multi-scale descriptors with scaling factors $(\frac{1}{\sqrt{2}}, 1, \sqrt{2})$ to improve performance. The three extracted descriptors are then sum-aggregated into a single vector followed by $\ell^2$-normalization.

\subsection{Local CNN model}\label{sec:localmodel}

For the local CNN retrieval model, we make use of the recently released Detect-to-Retrieve (D2R) \cite{TAZS2019} model which is publicly available\footnote{https://github.com/tensorflow/models/tree/master/research/delf}.

Compared to the first version of ``deep local features'' (DELF) \cite{NASWH2017}, a ResNet-50 \cite{HZRS2016} based Faster R-CNN \cite{RHGS2015} landmark detector trained on the newly released Google-Landmarks-Boxes-Dataset has been added, DELF has been re-trained on GLD, and a new aggregation method known as D2R-R-ASMK$^{*}$ has been proposed to boost retrieval accuracy. Given two images, local feature matching is performed followed by geometric verification with RANSAC \cite{FB1981} producing an integer-valued number representing the number of matching inliers. For simplicity, we refer to this as inlier score in the following.

\subsection{Step-1: Global Search}\label{sec:globalsearch}

This step is based entirely on global CNN models. To be specific, for each model we first match all (118K) test images vs. all (4.13M) training images by extracting the $\ell^2$-normalized descriptors and taking the dot product for cosine similarity. Then for each test image, we keep the top-10 closest neighbors in the training set along with their corresponding class labels. For this step, we make predictions using only the top-5 neighbors (the remaining 5 will be used in the next step) by accumulating class similarities in these 5 neighbors and then taking the class label with the highest score as prediction.

\subsection{Step-2: Local Search on Global Candidates}\label{sec:localsearch}

In this step, a second match is performed using the D2R model. For each test image, instead of matching it vs. all training images, only the top-10 neighbors from the previous step are used as candidates. The rest is same as Step-1 that we accumulate class inlier scores in the top-5 closest neighbors, and then predict based on the highest score.

\subsection{Step-3: Re-Ranking}\label{sec:rerank}

This re-ranking step aims at distinguishing real landmark images from distractors. Since we did not train a separate landmark/distractor classifier which allows us to leave empty prediction for distractors, the goal is to make landmark images rank higher in the submission by increasing their confidence scores. To this end, we proceed as follows. Given a full list of predictions ranked in descending order based on confidence scores, keep only the top-20000 predictions (the rest of the list will not be changed in this step and these predictions will simply be appended to the end after the process). Starting from the rank-1 test image, match vs. all other lower ranked images using the D2R model and filter out images with inlier scores smaller than a certain threshold (24 in our case). For those test images with inlier scores at least 24, re-rank them according to the inlier scores and append just below the rank-1 image along with their predicted class labels unchanged. This can be achieved by, for example, setting the new confidence score as $c_1 - k\epsilon$, where $c_1$ denotes the confidence score of the rank-1 image, $k$ is a non-negative integer (one for each), and $\epsilon$ is a small number so as to make sure the resulting number is greater than the confidence score of the previous rank-2 test image. Then we move on to the previous rank-2 image and match vs. other test images that are not already in the updated list, and continue this fashion for all top-N test images ($N < 1000$ to be specified) for two rounds of matches.

\section{Results}\label{sec:results}

We now present the results corresponding to the models and steps described above. For the competition, the public leaderboard is calculated with approximately $35\%$ of the test data whereas the private leaderboard, which determines the final rankings, is based on the other $65\%$.

We begin with Step-1 (Section \ref{sec:globalsearch}) results using global CNN models. The first five rows in Table \ref{tab:GlobalDescriptors} correspond to the five backbone networks from Section \ref{sec:globalmodels}, whereas DIR \cite{GARL2016, GARL2017} stands for the pre-trained ``deep image retrieval'' model that is publicly available\footnote{ https://europe.naverlabs.com/Research/Computer-Vision/Learning-Visual-Representations/Deep-Image-Retrieval/}, and ``AUW $\rightarrow$ Concatenate-6'' means first reducing the dimension of each of the six descriptors to 2048-D by AUW followed by concatenation. In the last two rows of Table \ref{tab:GlobalDescriptors}, ``Aggregate-7 (top-k)'' means we accumulate class similarities using the top-k closest neighbors of the above seven descriptors to make predictions. It turns out top-3 performs better on the private leaderboard, but we used top-5 for the next steps in the competition as it had a higher score on the public leaderboard.

\begin{table}[htb]
\begin{center}
\begin{tabular}{|c|c|c|}
\hline
Backbone/Method & Private & Public\\
\hline
\hline
ResNet-101 & 0.21752 & 0.18803\\
ResNeXt-101 (64$\times$4d) & 0.17753 & 0.14642\\
SE-ResNet-101 & 0.21282 & 0.18978\\
SE-ResNeXt-101 (32$\times$4d) & 0.19093 & 0.16340\\
SENet-154 & 0.19650 & 0.16865\\
DIR & 0.21360 & 0.17632\\
AUW $\rightarrow$ Concatenate-6 & 0.20137 & 0.16535\\
Aggregate-7 (top-5) & 0.25138 & \textbf{0.21534}\\
Aggregate-7 (top-3) & \textbf{0.26735} & 0.21178\\
\hline
\end{tabular}
\end{center}
\caption{Leaderboard performance for Global CNN descriptors.}
\label{tab:GlobalDescriptors}
\end{table}

Table \ref{tab:StepbyStep} contains the whole progress as the competition goes on. The first row ($\#(1)$) is just the best result from Table \ref{tab:GlobalDescriptors}. For the second row ($\#(2)$), we trained a simple linear SVM using approximately 20K of the images from the clean version of last year's training set as positives and 10K images from the bottom of the (lower ranked) predictions as negatives to filter out potential distractors by updating the confidence score as ``score += (output - threshold)'' if output is less than threshold, where we set threshold to be 0.55. This was our earlier attempt to deal with distractors before coming up with Step-3. For Step-3, we use $N = 330$ for $\#(3)$ and $N = 550$ for $\#(5)$ since $\#(4)$ achieved better performance than $\#(2)$.

In $\#(6)$, we noticed some of the top-ranked predictions from the submission of $\#(5)$ have same confidence scores as a result of Step-3 since the inlier scores are integer-valued, hence we made a small modification using the submission of $\#(1)$ by, for each test image, changing its confidence score to $c_5$ += $c_1/1000$, where $c_5$ denotes the confidence score from $\#(5)$ and $c_1$ denotes the corresponding confidence score from $\#(1)$. In $\#(7)$, by merge we mean taking the top-ranked (newly updated) predictions from $\#(6)$ and $\#(3)$ and merge them in an alternating fashion starting from the rank-1 prediction of $\#(6)$, which leads to our final submission.

Finally, the last row ($\#(8)$) corresponds to a submission which we unfortunately did not select. It was obtained by training an additional NN with 1 hidden layer using descriptors (of all training images) extracted from Step-1 as input followed by re-ranking steps and merging with the submission of $\#$(7). Due to its relatively lower score on the public leaderboard, we decided not to select it for the final submission because we thought the NN approach potentially added some wrong predictions/distractors with high confidence scores.

\begin{table}[htb]
\begin{center}
\begin{tabular}{|c|c|c|c|}
\hline
$\#$ & Method & Private & Public\\
\hline
\hline
(1) & Step-1 & 0.25138 & 0.21534\\
(2) & Step-1 $\rightarrow$ SVM & 0.29301 & 0.24759\\
(3) & Step-1 $\rightarrow$ SVM $\rightarrow$ Step-3 & 0.31098 & 0.27741\\
(4) & Step-1 $\rightarrow$ Step-2 & 0.31870 & 0.26782\\
(5) & Step-1 $\rightarrow$ Step-2 $\rightarrow$ Step-3 & 0.36767 & 0.31593\\
(6) & Modify (5) using (1) & 0.36787 & 0.31626\\
(7) & Merge (6) and (3) & \textbf{0.37606} & \textbf{0.32101}\\
\hline
\hline
(8) & Merge (7) and (NN $\rightarrow$ SVM $\rightarrow$ Step-3) & \textbf{0.37936} & 0.32100\\
\hline
\end{tabular}
\end{center}
\caption{Leaderboard performance step-by-step.}
\label{tab:StepbyStep}
\end{table}

\section{Conclusion}\label{sec:conclu}

In this paper, we presented a detailed solution of our approach for the Google Landmark Recognition 2019 Challenge which involves both global and local CNN retrieval models as well as several steps that lead to the final submission. In our opinion, we think the most important part of the solution is the re-ranking step as described in Section \ref{sec:rerank} which boosted our score to a competitive level.

{\small
\bibliographystyle{ieee}
\bibliography{bib}
}

\end{document}